\documentclass[runningheads]{llncs}

\usepackage{eccv}

\usepackage{eccvabbrv}

\usepackage{graphicx}
\usepackage{booktabs}

\usepackage[accsupp]{axessibility}  

\usepackage[pagebackref,breaklinks,colorlinks,citecolor=eccvblue]{hyperref}

\usepackage{orcidlink}
\usepackage{graphicx}
\usepackage{tikz}
\usepackage{comment}
\usepackage{amsmath,amssymb} 
\usepackage{color}
\usepackage{algorithm}
\usepackage{algorithmic}

\usepackage{microtype}

\usepackage{float}
\usepackage{graphicx}
\usepackage{booktabs} 
\usepackage{amsmath}
\usepackage{amssymb}
\usepackage{amsmath,amssymb}
\usepackage{mathrsfs}
\usepackage[numbers,compress]{natbib}

\global\long\def\vct#1{\boldsymbol{#1}}
\global\long\def\mat#1{\boldsymbol{#1}}

\global\long\def\opmat{\mbox{mat}}
\global\long\def\opdiag{\mbox{diag}}

\global\long\def\norm#1{\lVert#1\rVert}

\global\long\def\R{\mathbb{R}}

\global\long\def\vq{\boldsymbol{q}}

\global\long\def\vu{\boldsymbol{u}}
\global\long\def\vv{\boldsymbol{v}}

\global\long\def\vx{\boldsymbol{x}}
\global\long\def\vy{\boldsymbol{y}}
\global\long\def\vz{\boldsymbol{z}}

\global\long\def\vtheta{\boldsymbol{\theta}}

\global\long\def\mA{\boldsymbol{A}}

\global\long\def\mC{\boldsymbol{C}}

\global\long\def\mI{\boldsymbol{I}}

\global\long\def\mM{\boldsymbol{M}}

\global\long\def\mP{\boldsymbol{P}}

\global\long\def\mR{\boldsymbol{R}}

\global\long\def\mW{\boldsymbol{W}}

\global\long\def\a{\alpha}

\global\long\def\T{\top}

\global\long\def\vu{\vct{u}}

\begin{document}

\title{Differentiable Proximal Graph Matching} 


\author{Haoru Tan \and
Chuang Wang \and
Xu-Yao Zhang \and
Cheng-Lin Liu
}

\authorrunning{F.~Author et al.}

\institute{Institute of Automation, Chinese Academy of Sciences, Beijing, China \and
University of Chinese Academy of Sciences, Beijing, China\\
\email{\{tanhaoru2018,wangchuang,chenglin.liu\}@ia.ac.cn}}

\maketitle

\begin{abstract}
  Graph matching is a fundamental tool in computer vision and pattern recognition. In this paper, we introduce an algorithm for graph matching based on the proximal operator, referred to as differentiable proximal graph matching (DPGM). Specifically, we relax and decompose the quadratic assignment problem for  the graph matching into a sequence of convex optimization problems. The whole algorithm can be considered as a differentiable map from the graph affinity matrix to the prediction of node correspondence. Therefore, the proposed method is able to be organically integrated into an end-to-end deep learning framework to jointly learn both the deep feature representation and the graph affinity matrix. In addition, we provide a theoretical guarantee to ensure the proposed method converges to a stable point with a reasonable number of iterations. Numerical experiments show that PGM outperforms existing graph matching algorithms on diverse datasets such as synthetic data, CMU House. Meanwhile, PGM can fully harness the capability of deep feature extractor and achieves state-of-art performance on PASCAL VOC keypoints.
  \end{abstract}

\section{Introduction}
Graph matching (GM) is ubiquitous in computer vision and pattern recognition to establish correspondence between two sets of visual features. 
Applications ranges from 
object recognition \cite{graph_rec1,graph_rec2}, tracking \cite{graph_track1,graph_track2,graph_track3} and shape matching \cite{graph_shape1,graph_shape2,graph_shape3}, as well as computational biology \cite{graph_bio1}.
People, who deploy GM to practical applications,  encounter two major challenges. First,  graph matching algorithm is often hard to reach satisfiable performance due to its intrinsic nature of NP-hardness. 
Second,
one has to take considerable efforts to find a good feature representation to construct graph affinity matrix in the GM objective. 

Traditionally, people consider developing matching algorithm and seeking feature extractor as two separated problems. GM is commonly formulated as a quadratic assignment optimization problem \cite{koopman,lawlers}. 
Existing algorithms deploy different relaxation strategies, for example,  spectral methods \cite{sp1,sp2,sp3,sp4,sp5,sp6}, probabilistic approaches \cite{relax1,relax2,relax3,relax4,relax5}, or directly work on the integer optimization problem \cite{IPFP}.
Meanwhile, the features fed into the GM problem are majorly hand-crafted, {\em e.g.,}  using point locations in an image, or local pixel features, or, in recent works, got from a pretrained DNN feature extractor \cite{deep_feature1}\cite{deep_feature2}.


More recently, several works attempted to consider the two tasks of solving the GM problem and learning the feature representation as a joint problem
\cite{lgm, gmn, pia}.
In \cite{gmn}, the authors proposed a deep neural network model to simultaneously learn a feature extractor and an affinity matrix for GM.
The core idea is following: The GM solver, of which the major part is the spectrum method for computing the leading eigenvector of the affinity matrix, can be considered as a differentiable map from the node feature vector to the matching score. Thus, the GM solver can be appended after the feature extractor to build a new feedforward network. Then, the feature extractor and GM solver module are trained simultaneously by  the gradient backward propagation.  
In another work \cite{pia}, the authors deployed  the graph convolution network (GCN) as a solver for GM problem.
The feature extractor and the GCN module are jointly trained as well. 
Compared with \cite{gmn}, the GCN module contains additional learnable parameters in the message passing iteration, resulting better expressivity and reaching a state-of-art result in PASCAL VOC key-points matching problem.
On the other hand, the extra parameters can lead to  overfitting if training sample is small, and be more sensitive to noise.


In our work, we proposed a differentiable proximal graph matching (DPGM) module.
It can either be treated as an independent GM solver or be integrated into a deep learning framework to learn the node feature representation that is best for GM task.
Similar to \cite{gmn}, our DPGM module does not contains any learnable parameters, but compared to previous work \cite{gmn} in PASCAL VOC graph matching dataset, we achieve more than 10\% improvement. This result is even better than the current state-of-art \cite{pia}, which utilizes the power of GCN at a cost of training lots of extra parameters.  In addition,  we can also combine the proposed DPGM module  with the GCN and  further improve the matching performance.

To understand why our approach works well. We establish a series of theoretical and experimental analyses. We find that how to relax the original integer optimization problem of GM into a continuous  one is crucial for learning a better node feature representation. One  way is to simply relax discrete assignment variable into a continuous version while preserving the quadratic objective unchanged. This relaxation is commonly used in spectrum-based methods, and the follow-up learning-based method \cite{gmn}. Another method is to split the affinity matrix into two parts: a unitary potential and a pair-wise potential. This approach is a popular form of probability-based method, and is used in our DPGM module.
The two methods are equivalent if the decision variables take only discrete value 0 or 1, but has different performance in the continuous relaxation setting. The latter is significantly better than the former in the jointly training procedure.
Moreover, we provide an interpretation of the relaxation method used in our DPGM module from the mean-field variational inference. Based on this insight, we present a converges analysis of the proximal iterations, which justifies the validation of our proposed method. 

Finally, as comparison, we implements several other classical GM algorithms \cite{GAGM, RRWM, PSM} that can also be considered as differentiable modules.   
Among all of them, our DPGM consistently performs well in both classical separated training and joint training in various conditions, e.g. when the training and test samples have different labels, and when the number of training samples is drastically smaller than the number of test samples.

[TBA outline: The remaining part of this paper is organized as follows.]


\section{Problem Formulation}
In this section, we present a conventional formulation of the graph matching problem as a quadratic assignment problem.
\paragraph{Objective.} Let $\mathcal{G}_1 = (V_1, E_1)$ and $\mathcal{G}_2 = (V_2, E_2)$ be two undirected graphs
with an equal number of nodes, {\em i.e.,}$|V_1|=|V_2|=n$, where $V_i$ and $E_i$ for $ i = 1, 2$ represents nodes set and edges set of the graph $\mathcal{G}_i$ respectively.  For the case $|V_1|\neq|V_2|$, one can add additional auxiliary nodes into the graphs to meet the equal-number-of-nodes condition.
Graph matching aims to find a one-to-one correspondence map between the nodes of the two graphs such that a utility function of the corresponding nodes and edges is maximized.

\paragraph{Formulation.} Formally, graph matching is formulated as a quadratic assignment problem \cite{koopman,lawlers}. Let $\vx \in \{0,1\}^{n^2 \times 1}$ be an assignment vector such that if  node $i$ in $\mathcal{G}_1$ is mapped to node $j$ in $\mathcal{G}_2$, then $x_{ij}=1$, $x_{i j^\prime} = 0$ and 
$x_{i^\prime j}=0$ for all $i^\prime \in V_1 \setminus \{i\}$ and $j^\prime \in V_2 \setminus \{j\}$. 
The optimal assignment of graph matching is formulated as 
\begin{equation} \label{eq:opt}
\begin{aligned}
\max_{\vx  \in \{0,1\}^{n^2}} &\quad \vx^\T \mM \vx
\quad \quad \quad
\text{s.t.} &\quad \mC \vx = \mat{1},
\end{aligned}
\end{equation}
where the binary matrix $\mC \in \{0,1\}^{n^2 \times n^2}$ encodes $n^2$ linear constraints ensuring that
$\sum_{i\in V_1} x_{ij^\prime} = 1 $ and $\sum_{j \in V_2} x_{i^\prime j}=1$ for all $i^\prime \in V_1$ and $j^\prime \in V_2$. The non-negative affinity matrix $\mM \in \R_{+}^{n^2 \times n^2}$ defines the utility function of the graph matching problem. 

\paragraph{Decomposition and Construction of the Affinity Matrix $\mM$.}
The diagonal terms $\mM_{ii,jj}$ represents the reward if node $i$ in $\mathcal{G}_1$ is mapped to node $j$ in $\mathcal{G}_2$, and  the off-diagonal terms $\mM_{ii^\prime,j j^\prime}$ represents the reward if pair $(i,i^\prime)$ in $\mathcal{G}_1$  corresponds to pair $(j,j^\prime)$ in $\mathcal{G}_2$.  Based on this physical meaning,
it is convenient to decompose  the affinity matrix $\mM$ into two parts
\begin{equation} \label{eq:M}
\mM = \opdiag(\vu) + \mP,
\end{equation}
where $\vu$ is the diagonal vector of $\mM$, and the matrix $\mP$ contains all off-diagonal entries of $\mM$. Traditionally, the affinity matrix $\mM$ is defined manually based on concrete applications. For example, Koopman-Beckmann's formula \cite{koopman} constructs the affinity matrix by setting $\mM = \mA_1 \otimes \mA_2$, where $\mA_1$ and $\mA_2$ are adjacent matrices of $\mathcal{G}_1$ and $\mathcal{G}_2$ respectively. 
Intuitively, $[\mM]_{ii^\prime, jj^\prime} = 1$ if and only if both pairs of $(i, i^\prime)$
and $(j, j^\prime)$ are directly connected in $\mathcal{G}_1$ and $\mathcal{G}_2$ respectively, and   $[\mM]_{ii^\prime, jj^\prime} = 0$ otherwise. If only adjacent matrix is used, the diagonal terms $\vu$ are all zero.
In many computer vision applications, we usually have additional prior information on nodes correspondence, such as, locations of nodes in an image, local pixel property (e.g. SIFT). Those prior information forms a node feature vector.
A common way to construct node affinity vector is
 $[\vu]_{ij} = \exp \big( \vv_{1,i}^\T \mW  \vv_{2,j} \big)$,
where $\vv_{1,i}$ and $\vv_{2,j}$ are $i$th node's feature vector in $\mathcal{G}_1$ and $j$the node's feature vector in $\mathcal{G}_2$,  and $\mW$ is a weight matrix. We can either assign a $\mW$ mannually, e.g. setting
$\mW$ as an identity matrix, or learn a $\mW$ by using the data-driven approach.


\section{ Proximal Graph Matching} \label{sec:PGM}
\subsection{Continuous relaxation and entropy regulerizer}
We relax the original integer optimization problem  \eqref{eq:opt} to a continuous version:
\begin{equation} \label{eq:opt2}
\begin{aligned}
\min_{\vz \in \mR_{+}^{n^2 \times 1}} &\quad \mathcal{L}(\vz),  \quad \text{ with }  \mathcal{L}(\vz) =  - \vu^\T \vz - \vz^\T \mP \vz  +  h(\vz)\\
\text{s.t.} &\quad \mC \vz = \mat{1},
\end{aligned}
\end{equation}
with an entropy regularizer $h(\vz) =\lambda  \vz^\T \log(\vz)$. The scalar $\lambda$ is the regularizer parameter, and the matrix $\mC$ is the same as the one in \eqref{eq:opt}, which ensures that $\opmat{(\vz)} \in \R ^{n \times n }$ is a doubly stochastic matrix, {\em i.e.},$ \sum_{i \in V_1} [\vz]_{ij^{\prime}} = 1$ and $\sum_{j \in V_2} [\vz]_{i^{\prime} j}=1$ for all $i^{\prime} \in V_1$ and $j^{\prime} \in V_2$. 

The relaxed optimization problem \eqref{eq:opt2} relates to  \eqref{eq:opt} in the following way:
The entries of the assignment vector $\vx$ are either $0$ or $1$ in  \eqref{eq:opt}.
Thus, $x_{ij}^2 = x_{ij}$ for all $i\in V_1$ and $j\in V_2$. From the affinity decomposition \eqref{eq:M}, we have
\begin{equation} \label{eq:M-decop}
\vx^\T \mM \vx = \vu^\T \vx + \vx^\T \mP \vx
\end{equation}
Substituting \eqref{eq:M-decop} into \eqref{eq:opt}, and relaxing the binary assignment vector $\vx\in\{0,1\}^{n^2 \times 1}$ into a non-negative vector $\vz\in\R_{+}^{n^2 \times 1}$, we get \eqref{eq:opt2}.

Moreover, we note that the entropy regularized version \eqref{eq:opt2} can be considered as a mean-field variational inference of the original GM problem \eqref{eq:opt} to estimate the following Boltzmann-Gibbs distribution
\begin{equation}
p(\vx) = \frac{1}{Z(\lambda)} \exp\Big[- \frac{1}{\lambda} \Big( 
 \vu^\T \vx + \vx^\T \mP \vx
\Big) \Big]  \delta \Big( \mC \vx - \mat{1} \Big)
\end{equation}
under the independent approximation
$p(\vx) = \prod_{i \in V_1, j\in V_2} p(x_{ij})$. We explain the detailed connection between \eqref{eq:opt2} and the mean-field variational inference in the appendix.
Based on this insight, we have a few remarks.

\begin{remark} \label{rek:p}
The variable $z_{ij}$ in \eqref{eq:opt2} can be interpreted as the probability of $p(x_{ij}=1)$, where $p(x_{ij})$ is the marginal distribution of the Boltzmann-Gibbs distribution $p(\vx)$.  
When the temperature $\lambda$ tends to 0, $p(\vx)$ is the uniform distribution on the set of all global optimal solutions of \eqref{eq:opt}.
\end{remark}
\begin{remark}
The temperature parameter $\lambda$ controls the hardness of the optimization problem \eqref{eq:opt2}. 
In the high-temperate region $\lambda \gg 1$,  the entropy term $h(\vz)$ dominates \eqref{eq:opt2}. As $\lambda$ increases, the optimization becomes easier but the prediction becomes worse. In the limit $\lambda \to \infty$, it results a uniform prediction $\vz = \mat{1}/n$, where $n$ is the number of nodes.  In the low temperate region $\lambda \ll 0$,    the energy term, $\vu^\T\vz + \vz^\T \mP \vz$, dominates \eqref{eq:opt2}, but in this situation the optimization problem becomes hard with many local minimals in the energy landscape.
\end{remark}

%
%

\subsection{Proximal method}
Next, we review the general proximal method, and then apply it to solve the relaxed problem $\eqref{eq:opt2}$.
 
\paragraph{Introduction of the proximal method.} For a general optimization problem
\begin{equation} \label{eq:opt-gen}
\min_{\vz \in \mathcal{Z}} f(\vz) + h(\vz),
\end{equation}
where $\mathcal{Z}$ is a closed convex set, $h(\vz)$ is a convex function and $f(\vz)$ is a general function. Proximal method optimizes \eqref{eq:opt-gen} iteratively by solving a sequence of convex optimization problems. Let $\vz_t$ be the iterand at step $t$.  The update rule is
\begin{equation} \label{eq:prox}
\vz_{t+1} =   \mathop{\rm argmin}_{\vz \in \mathcal{Z}}\;
 \vz^\T \nabla f(\vz_t) + h(\vz) + \tfrac{1}{\beta_t}
D(\vz, \vz_t),
\end{equation}
where $\beta_t>0$ is a scalar parameter controlling the proximal weight, and $D(\cdot, \cdot)$ is a non-negative proximal function satisfying that  $D(\vx, \vy) = 0$ if and only if $\vx = \vy$. The proximal function $D(\cdot, \cdot)$  can be understood as a distance function for the geometry of $\mathcal{Z}$.
The two parts $f$ and $h$ in the objective function \eqref{eq:opt-gen} are treated separately in the proximal step \eqref{eq:prox}. The first part $f(\vz)$ contains complicated terms that we should use a first-order approximation $\vz^\T \nabla f(\vz_t)$, and the second part $h(\vz)$ is a simple convex term such that we can directly solve  \eqref{eq:prox} without any approximation on $h$.

\paragraph{Proximal Graph Matching}
We next use the proximal iteration \eqref{eq:prox} to solve \eqref{eq:opt2}.
The objective function in \eqref{eq:opt2} contains a concave energy term 
\begin{equation} \label{eq:def-f}
f(\vz)=-\vu^\T \vz - \vz^\T \mP \vz,
\end{equation} and a convex entropy term $h(z) = \lambda \vz^\T \log(\vz)$.
We choose 
the Kullback-Leibler divergence as the proximal function 
\begin{equation} \label{eq:KL}
D(\vx,\vy) =\vx^\T\log(\vx) - \vx^\T \log(\vy).
\end{equation} The KL divergence has a better description of the geometry of the variable $\vz$, as $z_{ij}$ can be interpreted as  the probability mapping node $i$ in $\mathcal{G}_1$ to node $j$ in $\mathcal{G}_2$. In addition, KL divergence is convex with 
$D(\vx, \vy) = 0 $ if and only if $\vx = \vy$. The domain $\mathcal{Z}$ in \eqref{eq:opt-gen} is a simplex described by $\mC\vz = \mat{1}$ and $\vz\geq 0$ same as \eqref{eq:opt2}.
Above all,  our proximal graph matching solves a sequence of convex optimization problems
\begin{equation} \label{eq:prox-GM}
\begin{aligned}
\vz_{t+1} =& \mathop{\rm argmin}_{\vz \in \R_{+}^{n^2\times 1}} - \vz^\T (\vu + \mP \vz_t)+ \big(\lambda + \tfrac{1}{\beta_t}\big) \vz^\T \log (\vz) - \tfrac{1}{\beta_t} \vz^\T \log(\vz_t)
\\
\text{s.t. }\quad & \mC \vz = \mat{1}.
\end{aligned}
\end{equation}

The convex optimization \eqref{eq:prox-GM} has an analytical solution
\begin{align} \label{eq:iter-1}
\widetilde{\vz}_{t+1} &=
\exp\Big[
\tfrac{\beta_t}{ 1+ \lambda \beta_t} ( \vu + \mP \vx_t) 
+ \tfrac{1}{1+\lambda \beta_t} \log(\vz_t)
\Big]
\\ \label{eq:iter-2}
\vz_{t+1} & = \text{Sinkhorn}(\widetilde{\vz}_{t+1}),
\end{align}
where $\text{Sinkhorn}(\vz)$ is the Sinkhorn-Knopp transform \cite{sinkhorn} that  maps 
a nonnegative matrix of size $n\times n$ to a doubly stochastic matrix.
Here, the input and output variables are $n^2$ vector. When use the Sinkhorn-Knopp transform, we reshape the input (output) as an $n\times n$ matrix ( $n^2$ vector) respectively. 
One can check that \eqref{eq:iter-1} outputs the minimum $\vz$ in \eqref{eq:prox-GM} 
 ignoring the constraint  $\mC \vz = \mat{1}$. 
The first-order derivative on the left-hand side of \eqref{eq:prox-GM} is 0 when $\vz = \widetilde{\vz}_{t+1}$ shown in \eqref{eq:iter-1}.
In addition, the exponential function in \eqref{eq:iter-1} ensures all entries of $\widetilde{\vz}_{t+1}$ are nonnegative.
At the second step \eqref{eq:iter-2}, the unconstrained solution $\widetilde{\vz}_{t+1}$ is mapped to 
$\vz_{t+1}$, a point in the valid simplex so that $\opmat(\vz_{t+1})$ is a $n\times n$ doubly stochastic matrix. In the appendix, we provide a detailed proof that $\vz_{t+1}$ is the solution of \eqref{eq:prox-GM}. Finally, we summarize our PGM algorithm in Algorithm~\ref{alg:PGM}.



\begin{algorithm*}[tb]
	\caption{:~~Proximal Graph Matching \label{alg:PGM}}
	\begin{algorithmic}
		\STATE {\bfseries Input:} graph $\mathcal{G}_1$ and $\mathcal{G}_2$ of node size $n$, parameter sequence $\{\beta_0, \beta_1, \beta_2, \beta_3....\}$, maximum iteration $T$.
		\STATE Compute~~ $\textbf{M}_{P}$, $\textbf{M}_{e}$.
		\STATE Initialize~~ $\textbf{X}_{0} \Longleftarrow {\rm Sinkhorn}(\textbf{M}_{P})$, $t \Longleftarrow 0$.
		\FOR{$t=0$ {\bfseries to} $T-1$}
		\FOR{$i=1$ {\bfseries to} $n^2$} 
		\STATE $\textbf{X}_{t+1}[i] \Longleftarrow \sum_{j \neq i}\textbf{M}_{e}[i, j]\textbf{X}_{t}[j]$~~~~~~~~~~~~~~~~~~ $\triangleright :$ \texttt{Message Passing }
		\STATE $\textbf{X}_{t+1}[i] \Longleftarrow \textbf{X}_{t+1}[i] +  \textbf{M}_{P}[i, i]$~~~~$\triangleright :$ \texttt{Local update with the unary potential }
		\STATE $\textbf{X}_{t+1}[i] \Longleftarrow \frac{\beta_t}{\beta_t+1}\textbf{X}_{t+1}[i] + \frac{1}{\beta_t+1}{\rm log}(\textbf{X}_{t}[i])$ ~~$\triangleright :$ \texttt{Local update with last iteration's outputs }
		\STATE $\textbf{X}_{t+1}[i] \leftarrow {\rm exp}(\textbf{X}_{t+1}[i])$~~~~~~~~~~~~~~~~~~~~~~~~~~~~ $\triangleright :$ \texttt{Nonlinear Transform}
		\ENDFOR
		\STATE $\textbf{X}_{t+1} \Longleftarrow {\rm Sinkhorn}(\textbf{X}_{t+1})$~~~~~~~~~~~~~~~~~~~~~~~~~ $\triangleright :$ \texttt{Bi-Stochastic Operation}
		\ENDFOR
		\STATE Output $\textbf{X}_{T}$
	\end{algorithmic}
\end{algorithm*}

\section{Convergence Analysis}
In this section, we show that under a mild assumption, the proposed DPGM method converges to a stationary point within a reasonable step. The main technique of the proof in this analysis follows  \cite{proximal_vi}, which studied variational inference problems using proximal-gradient method using a general divergence function. 
Before presenting the main proposition, we first show a technical lemma.
\begin{lemma} \label{lem:KL}
There exist a constant $\alpha >0$ such that for all $\vz_{t+1}$, $\vz_t$ generated by 
\eqref{eq:prox-GM}, we have
\begin{equation}
(\vz_{t+1} - \vz_{t})^\T \nabla_{\vz_{t+1}} D(\vz_{t+1}, \vz_{t}) \geq \alpha \norm{\vz_{t+1} - \vz_t}^2,
\end{equation}
where $D$ is the KL-divergence defined in \eqref{eq:KL}.
\end{lemma}
One can find the proofs of both of this lemma and the  proposition shown below in the supplementary materials.
Next, we show the main result of our convergence analysis. 
 
 \begin{proposition} \label{prop:conv}
 Let $L$ be the Lipschitz constant of the gradient $\nabla f(\vz)$, where $f(x)$ is defined in \eqref{eq:def-f}, and let $\alpha$ be the constant used in Lemma~\ref{lem:KL}.
 If we choose a constant step-size $\beta_t = \beta < 2\a/L$ for all $t\in \{0, 1, \ldots, T \} $, then
 \begin{equation}
 \begin{aligned}
 &\mathbb{E}_{t \sim \text{uniform} \{0, 1, \ldots, T \} }  \norm{\vz_{t+1} - \vz_t}^2
 \leq \frac{C_0}{ T(\alpha - L \beta /2)},
 \end{aligned}
 \end{equation} 
 where $C_0 = \mathcal{L}^\ast - \mathcal{L}(\vz_0)$ is the objective gap of \eqref{eq:opt2} between the initial guess
 $\mathcal{L}(\vz_0)$ and the optimal one $\mathcal{L}^\ast$.
 \end{proposition}
 
 On the Lipschitz constant $L$, if we construct $\mM$ by Koopman-Beckmann's form, then 
 $L=4 |E_1| \cdot |E_2|$, where $|E_1|$ and $|E_2|$ are the numbers of edges of the graph $\mathcal{G}_1$ and $\mathcal{G}_2$ respectively. This is due to the fact that 
 \begin{equation*}
 \norm{\nabla f(\vz) - \nabla f(\vy)} = \norm{\mM (\vz - \vy) } \leq \norm{\mM} \norm{(\vz - \vy)} ,
 \end{equation*}
 where $L = \norm{\mM} = \norm{A_1}_F \cdot \norm{A_2}_F = |E_1| \cdot |E_2|$.  
 
Proposition~\ref{prop:conv} guarantees that PGM converges to a stationary point within a reasonable number of iterations. 
In particular, the average difference $ \norm{\vz_{t+1} - \vz_t}^2$ of the iterand converges with a rate 
$\mathcal{O}(1/T)$.

\section{Integration of DPGM into Deep Neural Networks} \label{sec:learning}
In this section, we demonstrate how to integrate the proposed DPGM algorithm into a deep learning framework, so that one can utilize the data-driven approach to improve the performance
by learning the affinity matrix and fine-tuning the feature extractor.
The general idea follows \cite{gmn,pia}. The major difference between our approach and  \cite{gmn,pia} is that the graph matching layer uses DPGM module in our framework, whereas the counterpart in \cite{gmn} is the spectrum method, and \cite{pia} deployed a graph convolutional network as a graph matching solver. 


\subsection{Problem Setting of key points matching}
We take the problem of  supervised key points matching as an example to illustrate the learning framework. 
Suppose we have a set of a pair of images, where each image is marked some key points.
The task of key points matching aims to find corresponding points between the two images.
In the supervised case, we have a training set, where the ground truth of key points correspondence  are manually labeled. 
In practical situations, those image points pairs correspond to each other either by viewing the same 3d  physical object in different angles, or by the same semantic category of two different objects, e.g. tires of two different cars. 
In our model, we build the graph of key points associated with an image by 
 setting the key points as nodes. We then use
the Delaunay triangulation to generates edges,
{\em i.e.}, triangulation of the convex hull of the points on a plane in which every circumcircle of a triangle is an empty circle. 
 
\subsection{Feature extractor module}
The feature module takes an image pair $\mI_{p,1}$, $\mI_{p,2}$ and a set of key point positions $(w_{p,a,i}, h_{p,a,i})$ as the input, where $p$, $i$ are the image pair index, and point index respectively, and $a=1$ or $2$ indicates the first or second image in the image pair.  
The output is
a feature vector $\vv_{p,a, i}
$\begin{equation} \label{eq:feature}
\vv_{p,a, i} = \text{Feature}( \mI_{p,a,i}, (w_{p,a,i}, h_{p,a,i}); \vtheta ),
\end{equation} 
where $\vtheta$ represents all learnable parameters in the feature extractor.
The Feature map is implemented by a deep convolutional network.
Specifically, we use VGG16 pretrained with ImageNet as the backbone model.
Following  \cite{pia}, we extract the node feature from a coarse-grained layer via bilinear interpolation. We put additional details on the implementation in Section~\ref{sec:pas}.

\subsection{Optional Graph Convolution Module}
An optional graph convolution module can be attached after the feature extractor module.
The graph convolution module contains a sequence of graph convolution layer. 
Each layer transforms its input node features $\vv^{(\ell -1 )}_{p,i,a}$ into a new feature $\vv^{(\ell)}_{p,i,a}$, which encodes some local graph information
\begin{equation}
\vv^{(\ell)}_{p,i,a} = \text{GraphConv}^{(\ell)} ( \vv^{(\ell -1 )}_{p,i,a} ),
\end{equation}
where $\ell = 1,2,\ldots, L$ is the index of graph convolution layer, and $\vv^{(0)}_{p,i,a}=\vv_{p,i,a}$ is
the output from previous feature extractor module.
We use the same implementation of the graph convolution layer as \cite{pia}.
\subsection{Learnable affinity matrix}

We set the node affinity  $\vu_{p}$ of  $p$th image pair as
\begin{equation} \label{eq:weight}
[{\vu_{p}}]_{ij} = \exp \big( \vv_{p,1,i}^\T \mW  \vv_{p,2,j} \big),
\end{equation}
where $\mW$ is a parameter matrix that will be learned from the training data. 
The initial guess of $\mW$  is assigned as a diagonal matrix. 


The edge affinity matrix $\mP_{p}$ is generated  by
\begin{equation}
[\mP_{p}]_{ij,i^\prime j^\prime} = \exp \big( \vv_{p,1,i}^\T \vv_{p,2,j} \cdot  \vv_{p,1,i^\prime}^\T \vv_{p,2,j^\prime} \big), 
\end{equation}
where no learnable parameter is presented in this process.

\subsection{Differentiable Proximal Graph Matching Module}

The proposed DPGM module take the input of an affinity matrix $\mM_p= \opdiag (\vu_p) + \mP_p$, then  
output a continuous  version of an assignment matrix $\vz_p$ for any given $p$th image pair. 

In particular, we can consider the proximal graph matching algorithm described in Section~\ref{sec:PGM}  as a  map from the affinity matrix $\mM_p$ to 
the node assignment probability vector $\vz_p$
\begin{equation}
\vz_p = DPGM(\mM_p).
\end{equation}
It is straightforward to check that this map is differentiable, since both of \eqref{eq:iter-1} and \eqref{eq:iter-2} are differentiable with respective to 
its input iterand as well as $\vu$ and $\mP$.

The property of being differentiable is not unique for proximal graph matching algorithm. Many other GM algorithm can also be considered as a map from $\mM$ to $\vz$. Those algorithms are able to be integrated into this deep neural network framework as well. Previous work \cite{gmn}, which uses spectrum method as its GM solver, is a case.  Other conventional GM algorithms, for example, 
Re-weighted random walk matching (RRWM) \cite{RRWM} and 
Graduated assignment graph matching (GAGM)\cite{GAGM} both are differentiable.
We implement those algorithms as a differentiable module, and compared them with the proposed DPGM module. We found DPGM works best in various setting. Details are presented in  the next section. 
 
\subsection{Loss and Prediction}
In Remark~\ref{rek:p}, we mentioned that the output of the proximal graph matching module $[\vz_p]_{ij}$ can be interpreted as the probability  that node $i$ in $\mathcal{G}_1$ matches node $j$ in $\mathcal{G}_2$.

We define the final loss function  using the cross entropy:
\begin{equation}  \label{eq:loss}
\text{loss}  (\vz_p, \vz^\ast_p) = 
- \sum_{i \in V_1, j \in V_2}
\Big[ [\vz_p]_{ij}^\ast \log [\vz_p]_{ij} + (1 - [\vz_p^\ast]_{ij}) \log( 1- [\vz_p]_{ij} ) \Big]
\end{equation}
where  $\vz_{p}^\ast$ is the true assignment vector with $[\vz_{p}^\ast]_{ij}$ being 1 if node $i$ in $\mathcal{G}_1$ matches node $j$ in $\mathcal{G}_2$, and being $0$ otherwise.

Stacking all modules described from \eqref{eq:feature} to \eqref{eq:loss}, we get a trainable model, where the input is  an image pair $\mI_{1,p}, \mI_{2,p}$, and a set of key point locations $(w_{p,a,i}, h_{p,a,i})$, as well as the true assignment vector $\vz_p^\ast$, and the trainable parameters are the parameters $\vtheta$ in the feature extractor and the node weight matrix $\mW$ introduced in \eqref{eq:weight}.
Those parameters are optimized via the stochastic gradient descent where the gradients are automatically computed by a deep learning library.

\section{Experiments} \label{sec:exp}

In this section,  we conducted a series of experiments to test the performance of our proposed DPGM module.
We first use   the synthetic data and toy data with handcraft feature vectors to test its performance as an independent algorithm. 
Then, we test its learning ability when DPGM is integrated into a deep neural network with real-life data, e.g. PASCAL VOC keypoints dataset. Finally, we establish an ablation study and an additional robust test to understand  key performance factors underlying DPGM.

\subsection{Baseline Methods}
We list several existing methods as baselines in the experiments.

\begin{enumerate}
\item Graduated assignment graph matching (GAGM)\cite{GAGM} is developed by combining a softassign method with continuation methods, while paying close attention to sparsity. 

\item Re-weighted random walk matching (RRWM) \cite{RRWM} formulates graph matching as node selection on an association graph and then simulates random walks with reweighting jumps to solve the problem. 

\item Probabilistic spectral matching (PSM) \cite{PSM} introduces a probabilistic approach to solve the spectral graph matching formula.

\item Spectral matching (SM) introduces the spectral relaxation and then adopts spectral decomposition. 

\item Integer projected fixed point method (IPFP) \cite{IPFP} can take as input any initial solution and quickly converge to a solution obeying the initial discrete constraints. IPFP is not differentiable.

\item Because power iteration method is totally differentiable, \cite{gmn} proposed the graph matching network (GMN) which jointly combined spectral matching and deep feature extractors.

\item Permutation based cross-graph affinity (PCA) \cite{pia} learns the node-wise feature and the structure information by employing a graph convolutional network.
\end{enumerate}
We note that the first four methods are differentiable, and Method 4, spectrum matching (SM) is the base module of \cite{gmn}.
In Section \ref{sec:pas}, we  implement Methods 1--3 as a differentiable module and integrated them into the deep neural network framework. Their performance is compared with our DPGM  as well as two other learning-based methods \cite{gmn} and \cite{pia}.

\subsection{Handcrafted Feature Matching}

\subsubsection{Synthetic Graphs}
Similar to the experiments in \cite{ GAGM, RRWM, IPFP}, we test our DPGM on  
the synthetic Erdos-Renyi random graph data.

 The set the experiment as follows: We generate a reference graph $\mathcal{G}_1$ with $(n_{\text{in}} + n_{\text{out}})$ nodes, and randomly generated an edge between any node pair with a probability $p_{\text{edge}}$. 
 Each node $i$ in the reference graph $\mathcal{G}_1$ is associated with an feature vector $\textbf{q}_i$ uniformly drawn from $[0,1]^D$ with $D=20$.
  We duplicate the reference graph $\mathcal{G}_1$ as a new graph $\mathcal{G}_2$ preserving the same topology but we modify the feature vectors.
Then, we set $n_{\text{in}} $ nodes  as inliers and $n_{out}$ nodes as  outliers.
An inlier node means it has a matching in the other graph, while an outlier has no correspondence.
In $\mathcal{G}_2$, the feature vector of an inlier node is perturbed with an additive i.i.d. Gaussian noise with zero mean and variance $\sigma^2$ from its counterpart in $\mathcal{G}_1$, and for the outlier nodes, we generated a new i.i.d. feature vector following the same uniform distribution.

We construct the affinity matrix for all tested methods except PCA by setting the pairwise potential (edge affinity)  as $\mP_{ij^\prime, i^\prime j} = {\rm exp}(-|d_{ij}-d_{i^\prime j^\prime}|^2/2900 )$ if 
$(i,i^\prime) \in E_1$ and $(j,j^\prime)\in E_2$, and  $ \mP_{ij^\prime, i^\prime j} = 0$ otherwise, where $d_{ij}=\norm{\vq_i - \vq_j}$, and dropping the unary potential (node affinity). Permutation based intra-graph affinity (PCA) is based on the learnable graph neural network model. We train PCA with 2000 graph pairs for each task with the best parameter setting. Then, we fix the model, and transfer to other parameter setting. 

\begin{figure}[tbp] 
	\centering
	\includegraphics[width=0.98\linewidth]{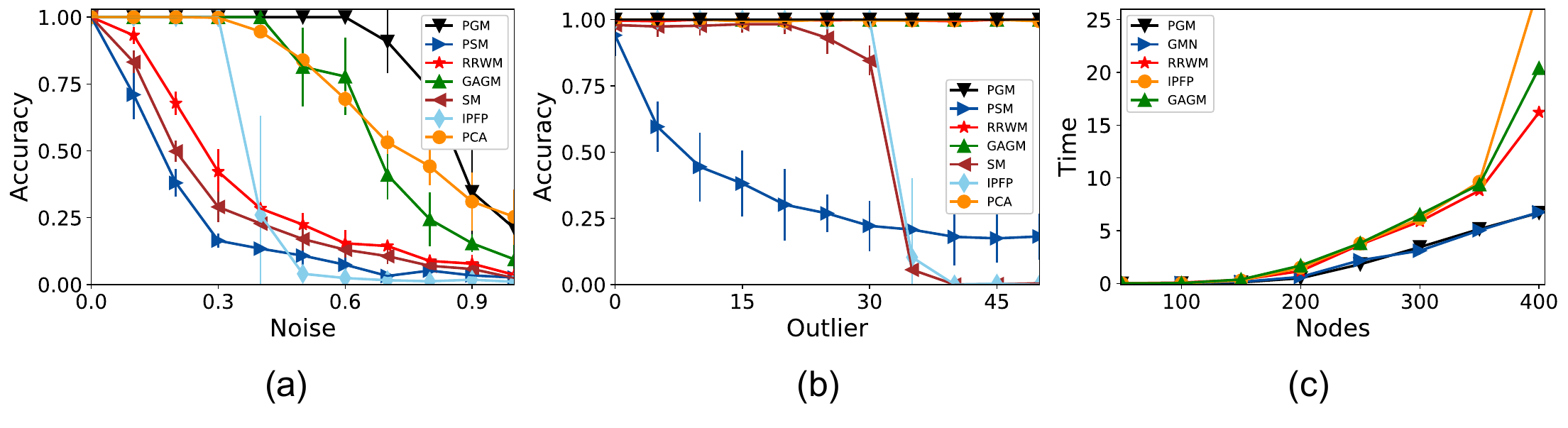}
	\caption{\label{ablation}Comparison of the matching accuracy, robustness of DPGM and other baselines on synthetic graph dataset. }
\end{figure}

\paragraph{Analysis robustness to noise.} We vary the noise level $\sigma$ from 0.0 to 2.0, with other parameters fixed as $n_{in} = 100$ , and $n_{out} = 0, p_{edge} = 0.7$. The results are shown in Figure \ref{ablation}(a).

\paragraph{Analysis the accuracy under different $n_{out}$ settings.} We change the number of outliers from $0$ to $50$ with $n_{in} = 35$, and $\sigma = 0.1, p_{edge} = 0.7$ being fixed. The results are presented in Figure \ref{ablation}(b).

In both experiments, our PGM outperforms all other baseline matching algorithms. We also note that the generalization performance of PCA \cite{pia} is also satisfactory, but it requires additional samples to train the model. On the other hand, the spectrum method, the base GM solver of GMN \cite{gmn} does not work well. This is our initial motivation for exploring different differentiable module to seeking a better learning based GM method.

\subsubsection{CMU House Sequence}

\begin{figure}[tb] 
	\centering 	
	\includegraphics[width=0.98\linewidth]{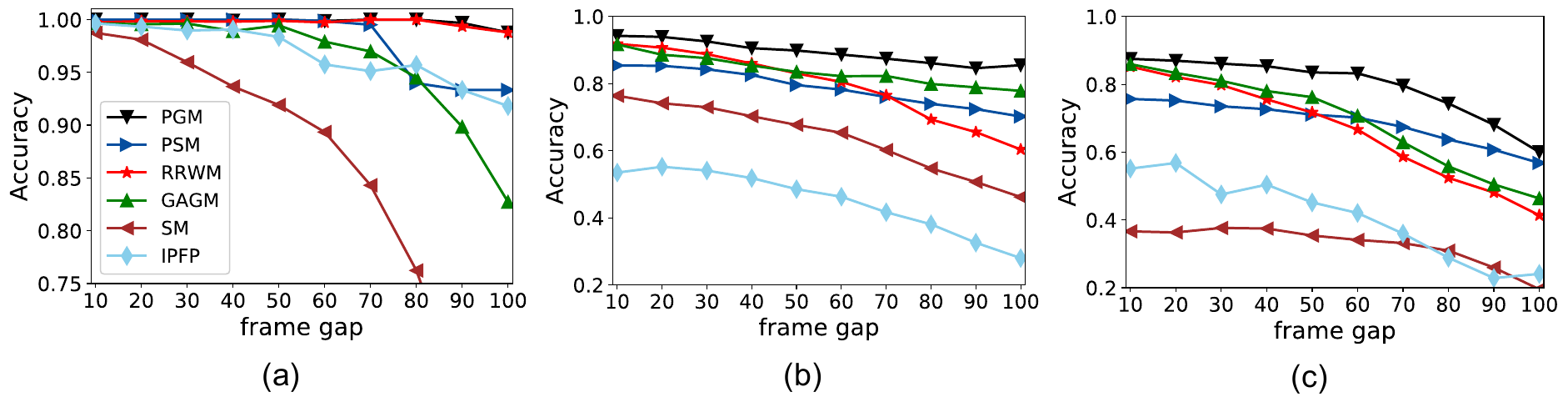}
	\caption{ \label{cmu}Comparison of the matching accuracy, robustness of PGM and other baselines on CMU house sequence dataset. }
	\label{hand_data2}
\end{figure}

The CMU House Sequence dataset contains totally 111 images of a house. Each image has 30 labeled landmarks. We build graphs by using these landmarks as nodes and adopting Delaunay triangulation to connect them. For all algorithms, affinity matrix is computed as $\textbf{M}_{ia, jb} = {\rm exp}(-|d_{ij}-d_{ab}|^2/2500 )$, where $d_{ij}$ means the distance between features $\textbf{q}_i$ and $\textbf{q}_j$ of two node from the same graph. We report the final average matching accuracy in Figure \ref{cmu}(a), \ref{cmu}(b) and \ref{cmu}(c) under different ratio settings of inliers-vs-outliers: 30:30, 25:30, 20:30. Our PGM achieves the best performance. Because the dataset is very small, grapg neural network based PCA cannot work here. Since the whole dataset too small to train PCA \cite{pia}, we does not add this method into comparison.

\subsection{Deep Feature Matching} \label{sec:pas}

The second group of experiments aims to test the compatibility of PGM and deep neural networks on the real-life PASCAL VOC dataset \cite{pascal,5459303}. The dataset has 20 classes of instances and  7,020 annotated images for training and 1,682 for testing. Berkeley's annotations provide labeled keypoint locations \cite{5459303}. 

\subsubsection{PASCAL VOC Keypoint}

The graph is built by setting those keypoints as nodes and then using Delaunay triangulation to introduce edges to connect landmarks. We use the feature from conv4-2 and conv5-1 layer of VGG-16 for all all situations \cite{gmn}\cite{pia}. Note that the width and height of feature map generated by VGG-16 is smaller than the given image, we cannot get the node features at precise location on the image level. Therefore, we use the bilinear interpolation to approximately recover the node feature of the given landmark pixel in the image. Similar technique is also used in \cite{pia,gmn}. 

We compared with our DPGM method with two recent learning-based methods GMN \cite{gmn}, and PCA \cite{pia}.  In addition, we also implement 
RRWM, GAGM and PSM as a differentiable module for comparison.

The results are listed in Table \ref{tb:pascal}. 
We classify the test models into two category. The first model in the first category
does not deploy any graph convolution network. Thus, there is no additional parameter other than those in the feature extractor. 
The second category attached a graph convolution network (GCN) after the feature extractor.  The experiments in both category shows that the model with DPGM module performs best. Furthermore, we find the PGM without GCN outperforms the PCA, where the latter involves much more parameters the the former.


\begin{table}[tb]
 \scriptsize
 \begin{center}
  \resizebox{\textwidth}{!}{\begin{tabular}{c|cccccccccccccccccccc|c}
    \hline
    method & aero &bike &bird &boat &bottle &bus &car &cat &chair &cow &table &dog &horse &mbike &person &plant &sheep &sofa &train &tv &mean \\ \hline \hline 
    
    GMN     &31.9  &47.2  &51.9  &40.8  &68.7  &72.2  &53.6  &52.8  &34.6  &48.6  &72.3  &47.7  &54.8  &51.0  &38.6  &75.1  &49.5  &45.0  &83.0  & 86.3  & 55.3  \\ 
    
    RRWM   & 11.3   &10.0 &16.7 &20.3 &12.2 &21.0 &18.9 &11.1 &11.8 &9.3 &22.9 &10.6 &10.8 &13.6 &9.0 &24.4 &15.4 &16.3 &29.5 &23.9 &15.9   \\  
    
    GAGM    &16.4 &18.9 &20.1 &27.2 &18.2 &25.4 &23.5 &15.7 &19.7 &14.1 &21.5 &15.7 &13.7 &19.6 &11.7 &36.7 &21.1 &18.0 &37.4 &36.7 &21.6  \\  
    
    PSM    &12.8 &12.6 &17.9 &24.2 &20.8 &37.9 &27.2 &12.8 &21.25 &14.9 &19.9 &13.6 &14.6 &16.1 &11.8 &24.7 &18.0 &21.3 &53.6 &42.91 &21.9  \\  
    
    PGM  &\textbf{45.6} &\textbf{60.6} &\textbf{52.79} &\textbf{50.3} &\textbf{76.3} &\textbf{81.1} &\textbf{69.6} &\textbf{63.5} &\textbf{42.7} &\textbf{59.3} &\textbf{83.5} &\textbf{58.1} &\textbf{66.4} &\textbf{59.8} &\textbf{43.6} &\textbf{81.8} &\textbf{64.9} &\textbf{56.3} &\textbf{85.5} &\textbf{89.4} &\textbf{64.6}          \\ \hline \hline 
    
    ${\rm \textbf{PCA}}^{*}$   &40.9  &55.0 & 65.8  &47.9  &76.9  &77.9  &63.5  &67.4  &33.7  &65.5  &63.6  &61.3  &68.9  &62.8  &44.9 & 77.5  &67.4  &57.5  &86.7  &90.9  &63.8          \\
    
    ${\rm \textbf{PGM}}^{*}$  &\textbf{49.7} &\textbf{64.8} &57.3 &\textbf{57.1} &\textbf{78.4} &\textbf{81.2} &\textbf{66.3} &\textbf{69.6} &\textbf{45.2} &63.4 &\textbf{64.2} &\textbf{63.5} &67.9 &62.0 &\textbf{46.8} &\textbf{85.5} &\textbf{70.7} &\textbf{58.1} &\textbf{88.9} &\textbf{89.9} &\textbf{66.58}         \\
    
    %
    %
		\end{tabular}}
	\end{center}
	
	\caption{Comparative matching results for PASCAL VOC Keypoint, where ${\rm \textbf{PGM}}^{*}$ uses the same VGG+GNN feature exactor \cite{pia} as ${\rm \textbf{PCA}}^{*}$,  while others use features extracted with the pure VGG model \cite{vgg}. }
	\label{tb:pascal}
	\vspace{-0.9cm}	
\end{table}

\section{Conclusion}

This paper has proposed a novel graph matching algorithm named as proximal graph matching (PGM). By introducing a proximal operator, PGM relax and decompose the original matching formula into a sequence of convex optimization objectives. By solving those convex sub-problems, PGM constructs a  differentiable map from the node and edge affinity matrix to the prediction of node correspondence. Comparison experiments show that PGM outperforms existing graph matching algorithms on diverse datasets such as synthetic data, CMU House and PASCAL VOC. which operation matters for deep feature matching is also thoroughly studied. We draw a conclusion from those experiments that, for deep feature matching, the node affinity is important. A good graph matching should help the deep model to be fully trained and introduce appropriate additional correction.

\bibliographystyle{splncs04}
\bibliography{main}

\end{document}